# Robust Feature Selection by Mutual Information Distributions

Marco Zaffalon
IDSIA
Galleria 2, 6928 Manno
Switzerland
zaffalon@idsia.ch

Marcus Hutter
IDSIA
Galleria 2, 6928 Manno
Switzerland
marcus@idsia.ch

## Abstract

Mutual information is widely used in artificial intelligence, in a descriptive way, to measure the stochastic dependence of discrete random variables. In order to address questions such as the reliability of the empirical value, one must consider sample-to-population inferential approaches. This paper deals with the distribution of mutual information, as obtained in a Bayesian framework by a second-order Dirichlet prior distribution. The exact analytical expression for the mean and an analytical approximation of the variance are reported. Asymptotic approximations of the distribution are proposed. The results are applied to the problem of selecting features for incremental learning and classification of the naive Bayes classifier. A fast, newly defined method is shown to outperform the traditional approach based on empirical mutual information on a number of real data sets. Finally, a theoretical development is reported that allows one to efficiently extend the above methods to incomplete samples in an easy and effective way.

## 1 Introduction

The *mutual information* $I$ (also called *cross entropy* or *information gain*) is a widely used information-theoretic measure for the stochastic dependency of discrete random variables [Kullback, 1968, Cover & Thomas, 1991, Soofi, 2000]. It is used, for instance, in learning *Bayesian nets* [Chow & Liu, 1968, Pearl, 1988, Buntine, 1996, Heckerman, 1998], where stochastically dependent nodes shall be connected; it is used to induce classification trees [Quinlan, 1993]. It is also used to select *features* for classification problems [Duda et al., 2001], i.e. to select a subset of variables by which to predict the *class* variable. This is done in the context of a *filter approach* that discards irrelevant features on the basis of low values of mutual information with the class [Lewis, 1992, Blum & Langley, 1997, Cheng et al., 2002].

The mutual information (see the definition in Section 2) can be computed if the joint chances $\pi_{ij}$ of two random variables $\imath$ and $\jmath$ are known. The usual procedure in the common case of unknown chances $\pi_{ij}$ is to use the *empirical probabilities* $\hat{\pi}_{ij}$ (i.e. the sample relative frequencies: $\frac{1}{n}n_{ij}$) as if they were precisely known chances. This is not always appropriate. Furthermore, the *empirical mutual information* $I(\hat{\boldsymbol{\pi}})$ does not carry information about the reliability of the estimate. In the Bayesian framework one can address these questions by using a (second order) prior distribution $p(\boldsymbol{\pi})$, which takes account of uncertainty about $\boldsymbol{\pi}$. From the prior $p(\boldsymbol{\pi})$ and the likelihood one can compute the posterior $p(\boldsymbol{\pi}|\mathbf{n})$, from which the distribution $p(I|\mathbf{n})$ of the mutual information can in principle be obtained.

This paper reports, in Section 2.1, the *exact* analytical mean of $I$ and an analytical $O(n^{-3})$-approximation of the variance. These are reliable and quickly computable expressions following from $p(I|\mathbf{n})$ when a *Dirichlet* prior is assumed over $\boldsymbol{\pi}$. Such results allow one to obtain analytical approximations of the distribution of $I$. We introduce asymptotic approximations of the distribution in Section 2.2, graphically showing that they are good also for small sample sizes.

The distribution of mutual information is then applied to feature selection. Section 3.1 proposes two new filters that use *credible intervals* to robustly estimate mutual information. The filters are empirically tested, in turn, by coupling them with the *naive Bayes classifier* to incrementally learn from and classify new data. On ten real data sets that we used, one of the two proposed filters outperforms the traditional filter: it almost always selects fewer attributes than the traditional one while always leading to equal or significantly better prediction accuracy of the classifier (Section 4).

The new filter is of the same order of computational complexity as the filter based on empirical mutual information, so that it appears to be a significant improvement for real applications.

The proved importance of the distribution of mutual information led us to extend the mentioned analytical work towards even more effective and applicable methods. Section 5.1 proposes improved analytical approximations for the tails of the distribution, which are often a critical point for asymptotic approximations. Section 5.2 allows the distribution of mutual information to be computed also from incomplete samples. Closed-form formulas are developed for the case of feature selection.

## 2 DISTRIBUTION OF MUTUAL INFORMATION

Consider two discrete random variables $\imath$ and $\jmath$ taking values in $\{1, ..., r\}$ and $\{1, ..., s\}$, respectively, and an i.i.d. random process with samples $(i,j) \in \{1, ..., r\} \times \{1, ..., s\}$ drawn with joint chances $\pi_{ij}$. An important measure of the stochastic dependence of $\imath$ and $\jmath$ is the mutual information:

$$I(\pi) = \sum_{i=1}^{r} \sum_{j=1}^{s} \pi_{ij} \log \frac{\pi_{ij}}{\pi_{i+}\pi_{+j}}, \quad (1)$$

where log denotes the natural logarithm and $\pi_{i+} = \sum_j \pi_{ij}$ and $\pi_{+j} = \sum_i \pi_{ij}$ are marginal chances. Often the chances $\pi_{ij}$ are unknown and only a sample is available with $n_{ij}$ outcomes of pair $(i,j)$. The empirical probability $\hat{\pi}_{ij} = \frac{n_{ij}}{n}$ may be used as a point estimate of $\pi_{ij}$, where $n = \sum_{ij} n_{ij}$ is the total sample size. This leads to an empirical estimate $I(\hat{\pi}) = \sum_{ij} \frac{n_{ij}}{n} \log \frac{n_{ij}n}{n_{i+}n_{+j}}$ for the mutual information.

Unfortunately, the point estimation $I(\hat{\pi})$ carries no information about its accuracy. In the Bayesian approach to this problem one assumes a prior (second order) probability density $p(\pi)$ for the unknown chances $\pi_{ij}$ on the probability simplex. From this one can compute the posterior distribution $p(\pi|\mathbf{n}) \propto p(\pi) \prod_{ij} \pi_{ij}^{n_{ij}}$ (the $n_{ij}$ are multinomially distributed) and define the posterior probability density of the mutual information:[1]

$$p(I|\mathbf{n}) = \int \delta(I(\pi) - I) p(\pi|\mathbf{n}) d^{rs}\pi. \quad (2)$$

---

[1] $I(\pi)$ denotes the mutual information for the specific chances $\pi$, whereas $I$ in the context above is just some non-negative real number. $I$ will also denote the mutual information *random variable* in the expectation $E[I]$ and variance $\text{Var}[I]$. Expectations are *always* w.r.t. to the posterior distribution $p(\pi|\mathbf{n})$.

[2] The $\delta(\cdot)$ distribution restricts the integral to $\pi$ for which $I(\pi) = I$. For large sample size $n \to \infty$, $p(\pi|\mathbf{n})$ is strongly peaked around $\pi = \hat{\pi}$ and $p(I|\mathbf{n})$ gets strongly peaked around the frequency estimate $I = I(\hat{\pi})$.

### 2.1 Results for $I$ under Dirichlet P(oste)riors

Many *non-informative* priors lead to a Dirichlet posterior distribution $p(\pi|\mathbf{n}) \propto \prod_{ij} \pi_{ij}^{n_{ij}-1}$ with interpretation $n_{ij} = n'_{ij} + n''_{ij}$, where $n'_{ij}$ are the number of samples $(i,j)$, and $n''_{ij}$ comprises prior information (1 for the uniform prior, $\frac{1}{2}$ for Jeffreys' prior, 0 for Haldane's prior, $\frac{1}{rs}$ for Perks' prior [Gelman et al., 1995]). In principle this allows the posterior density $p(I|\mathbf{n})$ of the mutual information to be computed.

We focus on the mean $E[I] = \int_0^\infty I p(I|\mathbf{n}) dI = \int I(\pi) p(\pi|\mathbf{n}) d^{rs}\pi$ and the variance $\text{Var}[I] = E[(I - E[I])^2]$. Eq. (3) reports the exact mean of the mutual information:

$$\begin{aligned} E[I] &= \frac{1}{n} \sum_{ij} n_{ij} [\psi(n_{ij}+1) - \psi(n_{i+}+1) \\ &\quad - \psi(n_{+j}+1) + \psi(n+1)], \end{aligned} \quad (3)$$

where $\psi$ is the $\psi$-function that for integer arguments is $\psi(n+1) = -\gamma + \sum_{k=1}^n \frac{1}{k} = \log n + O(\frac{1}{n})$, and $\gamma$ is Euler's constant. The approximate variance is given below:

$$\text{Var}[I] = \overbrace{\frac{K - J^2}{n+1}}^{O(n^{-1})} + \overbrace{\frac{M + (r-1)(s-1)\left(\frac{1}{2} - J\right) - Q}{(n+1)(n+2)}}^{O(n^{-2})} + O(n^{-3}) \quad (4)$$

where

$$K = \sum_{ij} \frac{n_{ij}}{n} \left(\log \frac{n_{ij}n}{n_{i+}n_{+j}}\right)^2,$$

$$J = \sum_{ij} \frac{n_{ij}}{n} \log \frac{n_{ij}n}{n_{i+}n_{+j}} = I(\hat{\pi}),$$

$$M = \sum_{ij} \left(\frac{1}{n_{ij}} - \frac{1}{n_{i+}} - \frac{1}{n_{+j}} + \frac{1}{n}\right) n_{ij} \log \frac{n_{ij}n}{n_{i+}n_{+j}},$$

$$Q = 1 - \sum_{ij} \frac{n_{ij}^2}{n_{i+}n_{+j}}.$$

The results are derived in [Hutter, 2001]. The result for the mean was also reported in [Wolpert & Wolf, 1995], Theorem 10.

---

[2] Since $0 \leq I(\pi) \leq I_{max}$ with sharp upper bound $I_{max} = \min\{\log r, \log s\}$, the integral may be restricted to $\int_0^{I_{max}}$, which shows that the domain of $p(I|\mathbf{n})$ is $[0, I_{max}]$.

We are not aware of similar analytical approximations for the variance. [Wolpert & Wolf, 1995] express the exact variance as an infinite sum, but this does not allow a straightforward systematic approximation to be obtained. [Kleiter, 1999] used heuristic numerical methods to estimate the mean and the variance. However, the heuristic estimates are incorrect, as it follows from the comparison with the analytical results provided here (see [Hutter, 2001]).

Let us consider two further points. First, the complexity to compute the above expressions is of the same order $O(rs)$ as for the empirical mutual information (1). All quantities needed to compute the mean and the variance involve double sums only, and the function $\psi$ can be pre-tabled.

Secondly, let us briefly consider the quality of the approximation of the variance. The expression for the exact variance has been Taylor-expanded in $\left(\frac{rs}{n}\right)$ to produce (4), so the relative error $\frac{\text{Var}[I]_{approx} - \text{Var}[I]_{exact}}{\text{Var}[I]_{exact}}$ of the approximation is of the order $\left(\frac{rs}{n}\right)^2$, if $\imath$ and $\jmath$ are dependent. In the opposite case, the $O\left(n^{-1}\right)$ term in the sum drops itself down to order $n^{-2}$ resulting in a reduced relative accuracy $O\left(\frac{rs}{n}\right)$ of (4). These results were confirmed by numerical experiments that we realized by Monte Carlo simulation to obtain "exact" values of the variance for representative choices of $\pi_{ij}$, $r$, $s$, and $n$.

## 2.2 APPROXIMATING THE DISTRIBUTION

Let us now consider approximating the overall distribution of mutual information based on the formulas for the mean and the variance given in Section 2.1. Fitting a normal distribution is an obvious possible choice, as the central limit theorem ensures that $p(I|\mathbf{n})$ converges to a Gaussian distribution with mean $E[I]$ and variance $\text{Var}[I]$. Since $I$ is non-negative, it is also worth considering the approximation of $p(I|\pi)$ by a Gamma (i.e., a scaled $\chi^2$). Even better, as $I$ can be normalized in order to be upper bounded by 1, the Beta distribution seems to be another natural candidate, being defined for variables in the $[0, 1]$ real interval. Of course the Gamma and the Beta are asymptotically correct, too.

We report a graphical comparison of the different approximations by focusing on the special case of binary random variables, and on three possible vectors of counts. Figure 1 compares the exact distribution of mutual information, computed via Monte Carlo simulation, with the approximating curves. The figure clearly shows that all the approximations are rather good, with a slight preference for the Beta approxima-tion. The curves tend to do worse for smaller sample sizes—as it is was expected—. Higher moments computed in [Hutter, 2001] may be used to improve the accuracy. A method to specifically improve the tail approximation is given in Section 5.1.

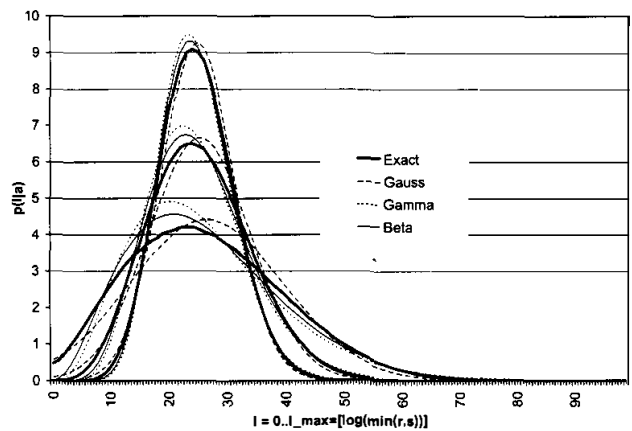

Figure 1: Distribution of mutual information for two binary random variables (The labeling of the horizontal axis is the percentage of I_max.) There are three groups of curves, for different choices of counts $(n_{11}, n_{12}, n_{21}, n_{22})$. The upper group is related to the vector $(40, 10, 20, 80)$, the intermediate one to the vector $(20, 5, 10, 40)$, and the lower group to $(8, 2, 4, 16)$. Each group shows the "exact" distribution and three approximating curves, based on the Gaussian, Gamma and Beta distributions.

## 3　FEATURE SELECTION

Classification is one of the most important techniques for knowledge discovery in databases [Duda et al., 2001]. A classifier is an algorithm that allocates new objects to one out of a finite set of previously defined groups (or *classes*) on the basis of observations on several characteristics of the objects, called *attributes* or *features*. Classifiers can be learnt from data alone, making explicit the knowledge that is hidden in raw data, and using this knowledge to make predictions about new data.

Feature selection is a basic step in the process of building classifiers [Blum & Langley, 1997, Dash & Liu, 1997, Liu & Motoda, 1998]. In fact, even if theoretically more features should provide one with better prediction accuracy (i.e., the relative number of correct predictions), in real cases it has been observed many times that this is not the case [Koller & Sahami, 1996]. This depends on the limited availability of data in real problems: successful models seem to be in good balance of model com-plexity and available information. In facts, feature selection tends to produce models that are simpler, clearer, computationally less expensive and, moreover, providing often better prediction accuracy. Two major approaches to feature selection are commonly used [John et al., 1994]: *filter* and *wrapper* models. The filter approach is a preprocessing step of the classification task. The wrapper model is computationally heavier, as it implements a search in the feature space.

### 3.1 THE PROPOSED FILTERS

From now on we focus our attention on the filter approach. We consider the well-known filter (F) that computes the empirical mutual information between features and the class, and discards low-valued features [Lewis, 1992]. This is an easy and effective approach that has gained popularity with time. Cheng reports that it is particularly well suited to jointly work with Bayesian network classifiers, an approach by which he won the *2001 international knowledge discovery competition* [Cheng et al., 2002]. The "Weka" data mining package implements it as a standard system tool (see [Witten & Frank, 1999], p. 294).

A problem with this filter is the variability of the empirical mutual information with the sample. This may allow wrong judgments of relevance to be made, as when features are selected by keeping those for which mutual information exceeds a fixed threshold $\varepsilon$. In order for the selection to be robust, we must have some guarantee about the actual value of mutual information.

We define two new filters. The *backward filter* (BF) discards an attribute if its value of mutual information with the class is less than or equal to $\varepsilon$ with given (high) probability $p$. The *forward filter* (FF) includes an attribute if the mutual information is greater than $\varepsilon$ with given (high) probability $p$. BF is a conservative filter, because it will only discard features after observing substantial evidence supporting their irrelevance. FF instead will tend to use fewer features, i.e. only those for which there is substantial evidence about them being useful in predicting the class.

The next sections present experimental comparisons of the new filters and the original filter F.

## 4 EXPERIMENTAL ANALYSES

For the following experiments we use the naive Bayes classifier [Duda & Hart, 1973]. This is a good classification model—despite its simplifying assumptions, see [Domingos & Pazzani, 1997]—, which often competes successfully with the state-of-the-art classifiers from the machine learning field, such as C4.5 [Quinlan, 1993]. The experiments focus on the incremental use of the naive Bayes classifier, a natural learning process when the data are available sequentially: the data set is read instance by instance; each time, the chosen filter selects a subset of attributes that the naive Bayes uses to classify the new instance; the naive Bayes then updates its knowledge by taking into consideration the new instance and its actual class. The incremental approach allows us to better highlight the different behaviors of the empirical filter (F) and those based on credible intervals on mutual information (BF and FF). In fact, for increasing sizes of the learning set the filters converge to the same behavior.

For each filter, we are interested in experimentally evaluating two quantities: for each instance of the data set, the average number of correct predictions (namely, the prediction accuracy) of the naive Bayes classifier up to such instance; and the average number of attributes used. By these quantities we can compare the filters and judge their effectiveness.

The implementation details for the following experiments include: using the Beta approximation (Section 2.2) to the distribution of mutual information; using the uniform prior for the naive Bayes classifier and all the filters; using natural logarithms everywhere; and setting the level $p$ of the posterior probability to 0.95. As far as $\varepsilon$ is concerned, we cannot set it to zero because the probability that two variables are independent ($I = 0$) is zero according to the inferential Bayesian approach. We can interpret the parameter $\varepsilon$ as a degree of dependency strength below which attributes are deemed irrelevant. We set $\varepsilon$ to 0.003, in the attempt of only discarding attributes with negligible impact on predictions. As we will see, such a low threshold can nevertheless bring to discard many attributes.

### 4.1 DATA SETS

Table 1 lists the 10 data sets used in the experiments. These are real data sets on a number of different domains. For example, Shuttle-small reports data on diagnosing failures of the space shuttle; Lymphography and Hypothyroid are medical data sets; Spam is a body of e-mails that can be spam or non-spam; etc.

The data sets presenting non-nominal features have been pre-discretized by MLC++ [Kohavi et al., 1994], default options. This step may remove some attributes judging them as irrelevant, so the number of features in the table refers to the data sets after the possible discretization. The instances with missing values have been discarded, and the third column in the table refers to the data sets without missing values. Finally,

Table 1: Data sets used in the experiments, together with their number of features, of instances and the relative frequency of the majority class. All but the Spam data sets are available from the UCI repository of machine learning data sets [Murphy & Aha, 1995]. The Spam data set is described in [Androutsopoulos et al., 2000] and available from Androutsopoulos's web page.

| Name | # feat. | # inst. | maj. class |
|---|---|---|---|
| Australian | 36 | 690 | 0.555 |
| Chess | 36 | 3196 | 0.520 |
| Crx | 15 | 653 | 0.547 |
| German-org | 17 | 1000 | 0.700 |
| Hypothyroid | 23 | 2238 | 0.942 |
| Led24 | 24 | 3200 | 0.105 |
| Lymphography | 18 | 148 | 0.547 |
| Shuttle-small | 8 | 5800 | 0.787 |
| Spam | 21611 | 1101 | 0.563 |
| Vote | 16 | 435 | 0.614 |

Table 2: Average number of attributes selected by the filters on the entire data set, reported in the last three columns. The second column from left reports the original number of features. In all but one case, FF selected fewer features than F, sometimes much fewer; F usually selected much fewer features than BF, which was very conservative. Boldface names refer to data sets on which prediction accuracies where significantly different.

| Data set | # feat. | FF | F | BF |
|---|---|---|---|---|
| Australian | 36 | 32.6 | 34.3 | 35.9 |
| **Chess** | 36 | 12.6 | 18.1 | 26.1 |
| Crx | 15 | 11.9 | 13.2 | 15.0 |
| German-org | 17 | 5.1 | 8.8 | 15.2 |
| Hypothyroid | 23 | 4.8 | 8.4 | 17.1 |
| Led24 | 24 | 13.6 | 14.0 | 24.0 |
| Lymphography | 18 | 18.0 | 18.0 | 18.0 |
| Shuttle-small | 8 | 7.1 | 7.7 | 8.0 |
| **Spam** | 21611 | 123.1 | 822.0 | 13127.4 |
| Vote | 16 | 14.0 | 15.2 | 16.0 |

the instances have been randomly sorted before starting the experiments.

## 4.2 RESULTS

In short, the results show that FF outperforms the commonly used filter F, which in turn, outperforms the filter BF. FF leads either to the same prediction accuracy as F or to a better one, using substantially fewer attributes most of the times. The same holds for F versus BF.

In particular, we used the *two-tails paired t test* at level 0.05 to compare the prediction accuracies of the naive Bayes with different filters, in the first $k$ instances of the data set, for each $k$.

On eight data sets out of ten, both the differences between FF and F, and the differences between F and BF, were never statistically significant, despite the often-substantial different number of used attributes, as from Table 2.

The remaining cases are described by means of the following figures. Figure 2 shows that FF allowed the naive Bayes to significantly do better predictions than F for the greatest part of the Chess data set. The maximum difference in prediction accuracy is obtained at instance 422, where the accuracies are 0.889 and 0.832 for the cases FF and F, respectively. Figure 2 does not report the BF case, because there is no significant difference with the F curve. The good performance of FF was obtained using only about one third of the attributes (Table 2).

Figure 3 compares the accuracies on the Spam data set. The difference between the cases FF and F is significant in the range of instances 32–413, with a maximum at instance 59 where accuracies are 0.797 and 0.559 for FF and F, respectively. BF is significantly worse than F from instance 65 to the end. This excellent performance of FF is even more valuable considered the very low number of attributes selected for classification. In the Spam case, attributes are binary and correspond to the presence or absence of words in an e-mail and the goal is to decide whether or not the e-mail is spam. All the 21 611 words found in the body of e-mails were initially considered. FF shows that only an average of about 123 relevant words is needed to make good predictions. Worse predictions are made using F and BF, which select, on average, about 822 and 13127 words, respectively. Figure 4 shows the average number of excluded features for the three filters on the Spam data set. FF suddenly discards most of the features, and keeps the number of selected features almost constant over all the process. The remaining filters tend to such a number, with different speeds, after initially including many more features than FF.

In summary, the experimental evidence supports the strategy of only using the features that are reliably judged as carrying useful information to predict the class, provided that the judgment can be updated as soon as new observations are collected. FF almost always selects fewer features than F, leading to a prediction accuracy at least as good as the one F leads to. The comparison between F and BF is analogous, so FF appears to be the best filter and BF the worst. However, the conservative nature of BF might turn out to be successful when data are available in groups, making the sequential updating be not viable. In this case, it does not seem safe to take strong decisions of exclusion that have to be maintained for a number of new instances, unless there is substantial evidence against the relevance of an attribute.

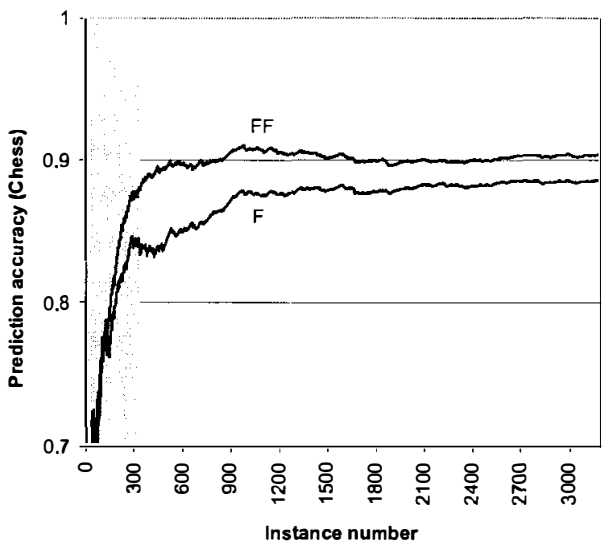

Figure 2: Comparison of the prediction accuracies of the naive Bayes with filters F and FF on the Chess data set. The gray area denotes differences that are not statistically significant.

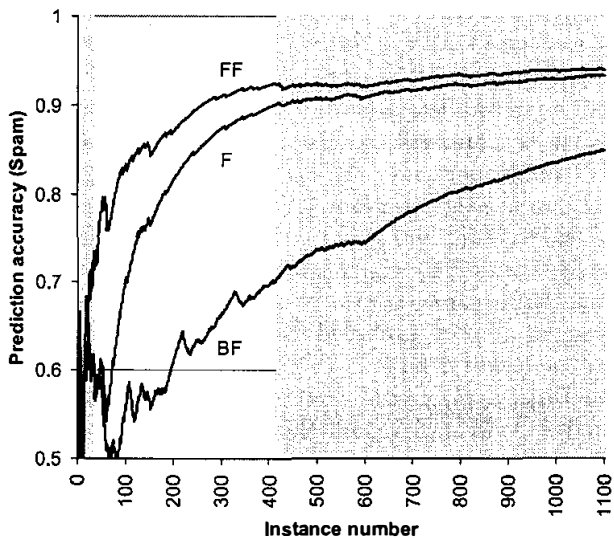

Figure 3: Prediction accuracies of the naive Bayes with filters F, FF and BF on the Spam data set. The differences between F and FF are significant in the range of observations 32–413. The differences between F and BF are significant from observations 65 to the end (this significance is not displayed in the picture).

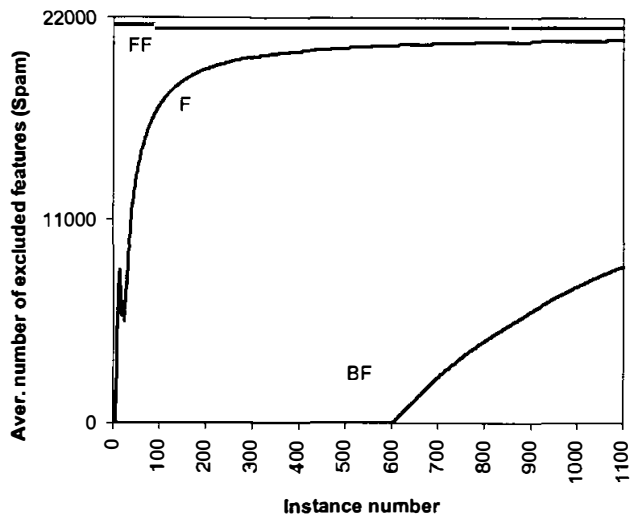

Figure 4: Average number of attributes excluded by the different filters on the Spam data set.

## 5 EXTENSIONS

### 5.1 TAILS APPROXIMATION

The expansion of $p(I|\mathbf{n})$ around the mean can be a poor estimate for extreme values $I \approx 0$ or $I \approx I_{max}$ and it is better to use tail approximations. The scaling behavior of $p(I|\mathbf{n})$ can be determined in the following way: $I(\pi)$ is small iff $\pi_{ij}$ describes near independent random variables $i$ and $j$. This suggests the reparameterization $\pi_{ij} = \tilde{\pi}_{i+}\tilde{\pi}_{+j} + \Delta_{ij}$ in the integral (2). Only small $\Delta$ can lead to small $I(\pi)$. Hence, for small $I$ we may expand $I(\pi)$ in $\Delta$ in expression (2). Correctly taking into account the constraints on $\Delta$, a scaling argument shows that $p(I|\mathbf{n}) \sim I^{\frac{1}{2}(r-1)(s-1)-1}$. Similarly we get the scaling behavior of $p(I|\mathbf{n})$ around $I \approx I_{max} = \min\{\log r, \log s\}$. $I(\pi)$ can be written as $H(i) - H(i|j)$, where $H$ is the entropy. Without loss of generality $r \leq s$. If the prior $p(\pi|\mathbf{n})$ converges to zero for $\pi_{ij} \to 0$ sufficiently rapid (which is the case for the Dirichlet for not too small $\mathbf{n}$), then $H(i)$ gives the dominant contribution when $I \to I_{max}$. The scaling behavior turns out to be $p(I_{max} - I_c|\mathbf{n}) \sim I_c^{\frac{r-3}{2}}$. These expressions including the proportionality constants in case of the Dirichlet distribution are derived in the journal version [Hutter & Zaffalon, 2002].

## 5.2 INCOMPLETE SAMPLES

In the following we generalize the setup to include the case of missing data, which often occurs in practice. For instance, observed instances often consist of several features plus class label, but some features may not be observed, i.e. if $i$ is a feature and $j$ a class label, from the pair $(i,j)$ only $j$ is observed. We extend the contingency table $n_{ij}$ to include $n_{?j}$, which counts the number of instances in which only the class $j$ is observed (= number of $(?,j)$ instances). It has been shown that using such partially observed instances can improve classification accuracy [Little & Rubin, 1987]. We make the common assumption that the missing-data mechanism is ignorable (missing at random and distinct) [Little & Rubin, 1987], i.e. the probability distribution of class labels $j$ of instances with missing feature $i$ is assumed to coincide with the marginal $\pi_{+j}$.

The probability of a specific data set $\mathbf{D}$ of size $N = n + n_{+?}$ with contingency table $\mathbf{N} = \{n_{ij}, n_{i?}\}$ given $\boldsymbol{\pi}$, hence, is $p(\mathbf{D}|\boldsymbol{\pi}, n, n_{+?}) = \prod_{ij} \pi_{ij}^{n_{ij}} \prod_i \pi_{i+}^{n_{i?}}$. Assuming a uniform prior $p(\boldsymbol{\pi}) \sim \delta(\pi_{++} - 1)$ Bayes' rule leads to the posterior $p(\boldsymbol{\pi}|\mathbf{N}) \sim \prod_{ij} \pi_{ij}^{n_{ij}} \prod_i \pi_{i+}^{n_{i?}} \delta(\pi_{++} - 1)$. The mean and variance of $I$ in leading order in $N^{-1}$ can be shown to be

$$E[\pi] = I(\hat{\boldsymbol{\pi}}) + O(N^{-1}),$$
$$\mathrm{Var}[I] = \frac{1}{N}[\tilde{K} - \tilde{J}^2/\tilde{Q} - \tilde{P}] + O(N^{-2}),$$

where

$$\hat{\pi}_{ij} = \frac{n_{i+} + n_{i?}}{N} \frac{n_{ij}}{n_{i+}}, \quad \rho_{ij} = N\frac{\hat{\pi}_{ij}^2}{n_{ij}}, \quad \rho_{i?} = N\frac{\hat{\pi}_{i+}^2}{n_{i?}},$$
$$\tilde{Q}_{i?} = \frac{\rho_{i?}}{\rho_{i?} + \rho_{i+}}, \quad \tilde{Q} = \sum_i \rho_{i+}\tilde{Q}_{i?},$$
$$\tilde{K} = \sum_{ij} \rho_{ij}\left(\log\frac{\hat{\pi}_{ij}}{\hat{\pi}_{i+}\hat{\pi}_{+j}}\right)^2, \quad \tilde{P} = \sum_i \frac{\tilde{J}_{i+}^2 Q_{i?}}{\rho_{i?}},$$
$$\tilde{J} = \sum_i \tilde{J}_{i+}\tilde{Q}_{i?}, \quad \tilde{J}_{i+} = \sum_j \rho_{ij}\log\frac{\hat{\pi}_{ij}}{\hat{\pi}_{i+}\hat{\pi}_{+j}}.$$

The derivation will be given in the journal version [Hutter & Zaffalon, 2002]. Note that for the complete case $n_{i?} = 0$, we have $\hat{\pi}_{ij} = \rho_{ij} = \frac{n_{ij}}{n}$, $\rho_{i?} = \infty$, $\tilde{Q}_{i?} = 1$, $\tilde{J} = J$, $\tilde{K} = K$, and $\tilde{P} = 0$, consistently with (4). Preliminary experiments confirm that FF outperforms F also when feature values are partially missing.

All expressions involve at most a double sum, hence the overall computation time is $O(rs)$. For the case of missing class labels, but no missing features, symmetrical formulas exist. In the general case of missing features and missing class labels estimates for $\hat{\pi}$ have to be obtained numerically, e.g. by the EM algorithm [Chen & Fienberg, 1974] in time $O(\# \cdot rs)$, where $\#$ is the number of iterations of EM. In [Hutter & Zaffalon, 2002] we derive a closed form expression for the covariance of $p(\boldsymbol{\pi}|\mathbf{N})$ and the variance of $I$ to leading order which can be evaluated in time $O(s^2(s+r))$. This is reasonably fast, if the number of classes is small, as is often the case in practice. Note that these expressions converge for $N \to \infty$ to the exact values. The missingness needs not to be small.

## 6 CONCLUSIONS

This paper presented ongoing research on the distribution of mutual information and its application to the important issue of feature selection. In the former case, we provide fast analytical formulations that are shown to approximate the distribution well also for small sample sizes. Extensions are presented that, on one side, allow improved approximations of the tails of the distribution to be obtained, and on the other, allow the distribution to be efficiently approximated also in the common case of incomplete samples. As far as feature selection is concerned, we empirically showed that a newly defined filter based on the distribution of mutual information outperforms the popular filter based on empirical mutual information. This result is obtained jointly with the naive Bayes classifier.

More broadly speaking, the presented results are important since reliable estimates of mutual information can significantly improve the quality of applications, as for the case of feature selection reported here. The significance of the results is also enforced by the many important models based of mutual information. Our results could be applied, for instance, to *robustly* infer classification trees. Bayesian networks can be inferred by using credible intervals for mutual information, as proposed by [Kleiter, 1999]. The well-known Chow and Liu's approach [Chow & Liu, 1968] to the inference of tree-networks might be extended to credible intervals (this could be done by joining results presented here and in past work [Zaffalon, 2001]).

Overall, the distribution of mutual information seems to be a basis on which reliable and effective uncertain models can be developed.

### Acknowledgements

Marcus Hutter was supported by SNF grant 2000-61847.00 to Jürgen Schmidhuber.